\documentclass[lettersize,journal]{IEEEtran}
\usepackage{amsmath,amsfonts}
\usepackage{algorithmic}
\usepackage{algorithm}
\usepackage{array}
\usepackage[caption=false,font=normalsize,labelfont=sf,textfont=sf]{subfig}
\usepackage{textcomp}
\usepackage{stfloats}
\usepackage{url}
\usepackage{verbatim}
\usepackage{graphicx}
\usepackage{cite}
\usepackage[colorlinks,linkcolor=red]{hyperref}

\usepackage{multirow}
\usepackage{amsthm}
\newtheorem{definition}{Definition}
\newtheorem{proposition}{Proposition}
\newtheorem{lemma}{Lemma}

\begin{document}

\title{Advancing Spiking Neural Networks towards Deep Residual Learning}

\author{Yifan Hu\textsuperscript{1}, Lei Deng\textsuperscript{1},~\IEEEmembership{Member,~IEEE,} Yujie Wu\textsuperscript{2}, Man Yao\textsuperscript{3}, Guoqi Li\textsuperscript{4,*},~\IEEEmembership{Member,~IEEE}
\thanks{\textsuperscript{1}Center for Brain-Inspired Computing Research, Department of Precision Instrument, Tsinghua University, Beijing, China. \textsuperscript{2}Technische Universität Graz Inffeldgasse, Graz, Austria. \textsuperscript{3}School of Automation Science and Engeneering, Xi'an Jiaotong University, Xi'an, Shaanxi, China. \textsuperscript{4}Institute of Automation, Chinese Academy of Sciences, Beijing, China. The corresponding author: Guoqi Li (E-mail:guoqi.li.ia.ac.cn).

This work was supported partially by National Key R\&D
Program of China (2018AAA0102600),  and Beijing Academy
of Artificial Intelligence (BAAI), and the Science and Technology Major Project of Guangzhou (202007030006), and Pengcheng Lab.
}
}

\markboth{FOR REVIEW}%
{Shell \MakeLowercase{\textit{et al.}}: A Sample Article Using IEEEtran.cls for IEEE Journals}

\IEEEpubid{0000--0000/00\$00.00~\copyright~2021 IEEE}

\maketitle

\begin{abstract}
Despite the rapid progress of neuromorphic computing, inadequate capacity and insufficient representation power of spiking neural networks (SNNs) severely restrict their application scope in practice. Residual learning and shortcuts have been evidenced as an important approach for training deep neural networks, but rarely did previous work assess their applicability to the characteristics of spike-based communication and spatiotemporal dynamics. In this paper, we first identify that this negligence leads to impeded information flow and the accompanying degradation problem in previous residual SNNs. To address this issue, we propose a novel SNN-oriented residual architecture termed MS-ResNet, which establishes membrane-based shortcut pathways, and further prove that the gradient norm equality can be achieved in MS-ResNet by introducing block dynamical isometry theory, which ensures the network can be well-behaved in a depth-insensitive way. Thus we are able to significantly extend the depth of directly trained SNNs, e.g., up to 482 layers on CIFAR-10 and 104 layers on ImageNet, without observing any slight degradation problem. To validate the effectiveness of MS-ResNet, experiments on both frame-based and neuromorphic datasets are conducted. MS-ResNet104 achieves a superior result of 76.02\% accuracy on ImageNet, which is the highest to our best knowledge in the domain of directly trained SNNs. Great energy efficiency is also observed, with an average of only one spike per neuron needed to classify an input sample. We believe our powerful and scalable models will provide a strong support for further exploration of SNNs.
\end{abstract}


\section{Introduction}
\IEEEPARstart{S}{piking} neural networks with unique features of rich neuronal dynamics and diverse coding schemes, represent a typical type of brain-inspired computing models. Different from traditional artificial neural networks (ANNs), SNNs are capable of encoding information in spatiotemporal dynamics and using asynchronous binary spiking activity for event-driven communication. Recent progress in neuromorphic computing has demonstrated their great potential in energy efficiency~\cite{roy2019towards,pei2019towards,mayr2019spinnaker,akopyan2015truenorth,davies2018loihi}. Theoretically, SNNs are at least as computationally powerful as ANNs and the universal approximation theorem also applies to SNNs~\cite{maass1997networks}. Therefore, it is not surprising that SNNs have been reported in various application tasks, such as image classification~\cite{Wu_Deng_Li_Zhu_Xie_Shi_2019}, object detection~\cite{Kim_Park_Na_Yoon_2020} and tracking~\cite{yang2019dashnet}, 
speech recognition~\cite{speech_recog}, light-flow estimation~\cite{lee2020spike}, and so forth. However, in practice, the lack of powerful SNN models seriously limits their capabilities for complex tasks. 

To obtain an SNN model, conversion from a pretrained ANN model and the surrogate gradient-based direct training are two mainstream approaches. The basic idea of the former is that the activation values in a ReLU-based ANN can be approximated by the average firing rates of an SNN under rate-coding scheme. After training an ANN with certain restrictions, it is feasible to convert the pretrained ANN into its spiking counterpart. In this way, the conversion methods are free of the dilemma caused by the non-differentiable spiking activation function and maintain the smallest gap with ANNs in terms of accuracy and get generalized to large-scale structures and datasets. Rueckauer et al.~\cite{rueckauer2017conversion} report the spiking versions of VGG-16 and GoogleNet. Hu, Tang and Pan~\cite{hu2018spiking} offer a compensation mechanism to reduce the discretization error, and obtain an accuracy of 72.75\% on ImageNet with spiking ResNet-50. In \cite{stockl2021optimized}, it is allowed for spikes to carry time-varying multi-bit information so that more closely the original activation function can be emulated and large EfficientNet models can be trained. However, the conversion routine also suffers from inherent defects. An accuracy gap will be caused by the constraints on ANN models and a long simulation with hundreds or thousands of timesteps is required to complete an inference, which leads to extra delay and energy consumption. Therefore, recently more works have begun to focus on the economical reduction of timesteps~\cite{Li_freelunch_2021,deng2021optimal}, but there still exists severe accuracy loss when the timesteps are only a few dozen.

\IEEEpubidadjcol

 The latter method is to utilize a surrogate gradient function, which constitutes a continuous relaxation of the non-smooth spiking activity to enable standard backpropagation through time (BPTT) for training an SNN from scratch. Lee et al.~\cite{lee2016training} treat the membrane potentials as differentiable signals and discontinuities as noise, and start to train SNNs directly from spike signals. Wu et al.~\cite{wu2018spatio} build an iterative LIF model and propose spatio-temporal backpropagation (STBP) based on an approximate derivative for spiking activity. The direct training algorithms exhibit a diversity in the form of gradient functions~\cite{neftci2019surrogate} and coding schemes~\cite{zhang2020spike,Zhou_Li_Chen_Chandrasekaran_Sanyal_2021}. These directly trained networks learn to encode information effectively and consequently need much fewer timesteps than the conversion ones, which can be particularly appealing for the implementation on power-efficient neuromorphic hardware. In addition, they are inherently more suited for processing spatio-temporal data from the emerging AER-based (address-event-representation) sensors to which the pretrained ANNs and converted SNNs are not applicable~\cite{hagenaars2021self,deng2020rethinking}. Unfortunately, one prominent problem of the directly trained SNNs lies in the limited scale of models. The depth of neural networks is surely crucial for their success, but earlier directly trained SNN works mainly focus on shallow structures and simple tasks such as MNIST and N-MNIST. Inspired from the representation power of deep ANNs, recent works have gradually evolved from fully-connected networks to convolutional networks and then to more advanced ResNet~\cite{Zheng_Wu_Deng_Hu_Li_2021,fang2021deep,deng2022temporal}. Since residual learning and shortcuts have been evidenced as an important way of deepening ANNs and widely adopted~\cite{He_2016_resnet,He_2016_identitymapping,Huang_2017_CVPR_densenet,10.5555/3298023.3298188_inception_resnet,10.5555/3295222.3295349_transformer}, it is natural to seek ways they can be applied in SNNs, for example Zheng et al. propose threshold-dependent batch normalization (TDBN) based on the framework of STBP~\cite{wu2018spatio} and first report the successful training of spiking ResNet-34 on ImageNet~\cite{Zheng_Wu_Deng_Hu_Li_2021}. 

However, rarely do previous works assess the applicability of residual learning to the intrinsic characteristics of spike-based communication scheme and spatiotemporal dynamics of SNNs. The direct adoption of well-developed ANN structures seems to have become a stereotype for SNN structure design, but we find that the plain transplantation of canonical ResNet, a way almost all previous works have adopted, does not work appropriately for the training of SNNs. The issue will manifest itself in our depth-analysis experiments as an accuracy drop on both the training and test sets as the network deepens, which is termed the degradation problem. As a result, building a powerful deep SNN model still remains arduous and fruitless, which severely hinders its utilization in various tasks, and an SNN-oriented residual block design is highly desirable for obtaining an SNN model with great scalability to pursue sufficient representation power by adequate capacity.

In this article, we first report the degradation problem when applying vanilla ResNet in training deep SNNs. The accuracy improvement trend of spiking PlainNet and vanilla ResNet with increasing depth stops at 14 and surprisingly only 20 layers, respectively. To address the degradation problem and fully unleash the potential of deep SNNs, we propose an SNN-oriented residual architecture, MS-ResNet, where the interblock $\text{LIF}(\cdot)$ is removed to construct a clean shortcut connection imposing identity mapping throughout the network. Then we analyze the superiority of the new structure from two perspectives, the forward propagation and the gradient norm, respectively. It is observed that MS-ResNet can avoid unavailing residual representations at inference, which refer to the case that the residual path of a block does not contribute to the overall capacity of the network. In addition, we prove that MS-ResNet can achieve gradient norm equality by the framework of block dynamical isometry~\cite{chen2020comprehensive} while the vanilla spiking ResNet cannot. Consequently, our model shows great scalability and can be extended at least to 482 layers on CIFAR-10 and 104 layers on ImageNet, without observing degradation problem. We also evaluate the effectiveness of MS-ResNet on various datasets and obtain accuracy results which are, to the best of our knowledge, the state-of-the-art in directly trained SNNs and competitive to the conversion methods but with extremely fewer timesteps. Along with the sparse spiking activity, great energy efficiency is also validated and it achieves up to 24x energy saving in processing event streams.

The rest of this article is organized as follows. In Section~\ref{sec:Preliminaries and Motivation}, we introduce the preliminaries of SNNs and our motivation originating from the degradation problem. In Section~\ref{sec:Spiking Residual Blocks}, our SNN-oriented residual block is proposed along with the analysis of the superiority. Benchmark experiments and energy efficiency estimation are provided in Section~\ref{sec:Experiments}. Finally, we discuss some interesting details in Section~\ref{sec:Discussion} and conclude in Section~\ref{sec:Conclusion}.

\section{Preliminaries and Motivation}
\label{sec:Preliminaries and Motivation}
\subsection{Preliminaries of SNNs}
The basic differences between an SNN and an ANN originate from their primary computing element, i.e., the neuron. In ANNs, a biological neuron is abstracted as an information aggregation unit with a nonlinear transformation. On the contrary, in SNNs, the membrane potential dynamics and the spiking communication scheme are more closely mimicked. Typically, there are several kinds of spiking neuron models, such as LIF model~\cite{abbott1999lapicque}, Izhikevich model~\cite{Izhikevich_model}, and Hodgkin-Huxley model~\cite{hodgkin1952quantitative}. At the level of large-scale neural networks, LIF models are widely adopted due to the concise form and low computational complexity. In this work, we select the iterative LIF model proposed by Wu et al.~\cite{Wu_Deng_Li_Zhu_Xie_Shi_2019} for SNN modelling, which can be formulated as
\begin{align}
    u_i^t&={\tau}_{mem}\cdot u_i^{t-1}+\sum_{j=1}^{n}w_{ij}o_j^t, \\
    o_i^t&=g\left(u_i^t\right)=\begin{cases}
    1,& \text{if}~u_i^t-V_{th}\geq 0\\
    0,&\text{otherwise}
    \end{cases},
    \label{Eqn:gate_function}
\end{align}
where $u_i^t$ is the membrane potential of the $i$-th neuron in a layer at timestep $t$, $\tau_{mem}$ is a decay factor for leakage, and the synaptic input is the weighted sum of output spikes from the previous layer. $g(\cdot)$ describes the firing activity controlled by the threshold $V_{th}$ and $u_i^t$ will be subsequently reset to $V_{reset}$ once a spike fires, i.e. \( o_i^{t-1}=1\). The surrogate gradient is defined as in~\cite{wu2018spatio}
\begin{equation}
    \frac{\partial o_i^t}{\partial   u_i^t}=\frac{1}{a}\text{sign}\left(\left|u_i^t-V_{th}\right|\leq \frac{a}{2}\right).
    \label{Eqn:Surrogate_func}
\end{equation}
The coefficient $a$ is introduced to ensure that the integral of the function is 1. In this way, BPTT can be carried out along with the autograd framework of mainstream deep learning libraries.

\begin{figure}
    \centering
    \includegraphics[width=0.9\linewidth]{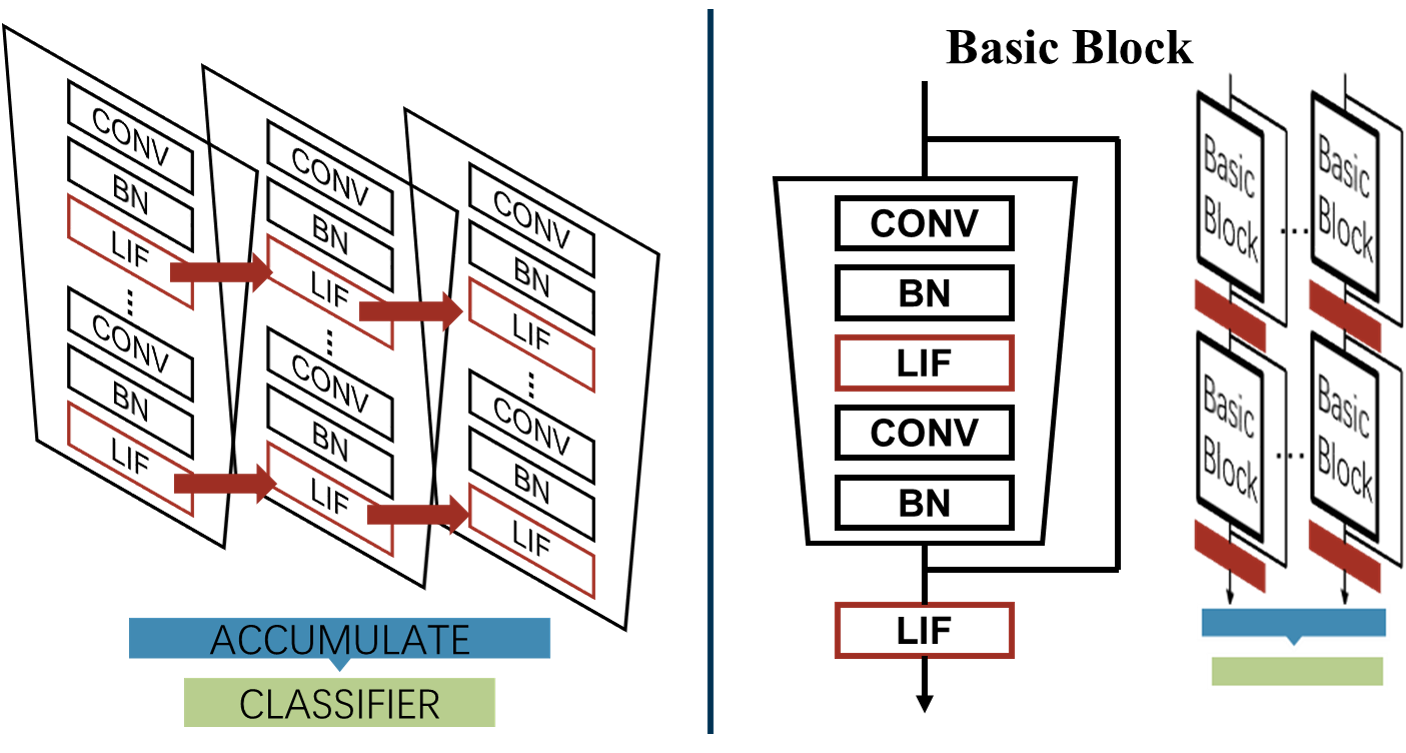}
    \caption{Spiking PlainNet and vanilla ResNet in SNNs.}
    \label{fig:vanilla-resnet}
\end{figure}

 For computer vision tasks, a stacking of \{Conv-BN-Nonlinearity\} is a universal architecture which follows the primary philosophy of VGG networks, and is referred to as PlainNet in this work. We adopt the recent TDBN technique \cite{Zheng_Wu_Deng_Hu_Li_2021} as the conventional BN in our spiking models and formulate it as
\begin{equation}
    u_i^t=\tau_{mem}u_i^{t-1}+\text{TDBN}\left(I_i^t,\mu_{c_i}, \sigma_{c_i}^2, V_{th}\right),
\end{equation}
where $\mu_{c_i},\sigma_{c_i}^2$ are channel-wise mean and variation values calculated per-dimension over a mini-batch of the sequential inputs $\{I_{i}^t=\sum_{j=1}^{n}w_{ij}o_j^t|t=1,...,T\}$.

A shortcut connection can be inserted into PlainNet and turns it into its residual counterpart, which can be written as
\begin{equation}
    \boldsymbol{o^{l+1}}=\text{LIF}\left(\mathcal{F}\left(\boldsymbol{o^{l};W^{l+1}}\right)+\boldsymbol{o^{l}}\right),
\end{equation}
where $\mathcal{F}(\cdot)$ represents the group of functions in a residual path with $\boldsymbol{W^{l+1}}$ as its parameters, $\boldsymbol{o^{l}}$ is the spike train output vector over a simulation period $T$, and $l$ represents the layer index. At the end of the model, a fully-connected layer counts the number of spikes from each neuron of the last feature map during the entire simulation and works as a classifier to make the final decision. 
The direct transplantation of a residual block from non-spiking ResNet is shown in Figure~\ref{fig:vanilla-resnet}, which is a conventional practice in previous works~\cite{Zheng_Wu_Deng_Hu_Li_2021,hu2018spiking}.

\subsection{The Degradation Problem in SNNs}

The most straightforward way of training higher quality models is to increase their size, especially given the availability of a large amount of labeled training data~\cite{Szegedy_2015_CVPR}. However, directly deepening the network has never seemed to be a trustworthy approach toward more satisfying accuracy in the field of SNNs.

We would like to explore how the existing SNN-oriented BN technique and shortcut connections contribute to the scalability of spiking models, so an experiment on CIFAR-10~\cite{krizhevsky2009learning} with the depth as the only variable is conducted. It should be noted that our focus is on the response of a network to its depth and the potential degradation problem rather than obtaining state-of-the-art results, so we use deep but relatively narrow architectures as in Table~\ref{tab:narrownet}.

Table~\ref{tab:depth_analysis} shows the results of the depth analysis. The accuracy of PlainNet begins to drop at the depth of 14 layers, and surprisingly the adoption of shortcut connections will just shift the peak to 20 layers. Despite a slight slope after the peak, severe accuracy decrease still occurs in spiking ResNet when it reaches 56 layers. The degradation problem does exist in spite of the adoption of TDBN and shortcut connections, indicating that the direct transplantation of ResNet to SNNs does not work effectively and making building sufficiently deep and powerful SNN models a nontrivial task. 

Interestingly, we have noticed that the degradation problem could be alleviated within the depth of 56 layers when we remove the spiking activation functions \(\text{LIF}(\cdot)\) between residual blocks and maintain those in the residual paths as nonlinearity. Based on this observation, we mainly identify the crux of degradation as the interblock \(\text{LIF}(\cdot)\), and will provide further insights along with our proposed architecture in the next section.

\begin{table}
\caption{The Structure for Depth Analysis on CIFAR-10\label{tab:narrownet}}
\centering
\begin{tabular}{lccc}
\hline
Layers          & $1+2n$  & $2n$    & $2n$  \\
\hline
Output Size & 32x32 & 16x16 & 8x8 \\ \hline
Channels         & 16    & 32    & 64  \\
\hline
\end{tabular}
\end{table}

\begin{table}[!t]
    \caption{Depth Analysis on CIFAR-10.}
    \label{tab:depth_analysis}
    \centering
    \begin{tabular}{lcccccc}
    \hline
    Depth       & 8     & 14    & 20    & 32    & 44    & 56    \\
    \hline
    PlainNet    & 81.1 & \textbf{85.2}  & 83.7 & 76.4 & 59.5  & N/A    \\
    Vanilla ResNet & 81.5  & 85.9 & \textbf{86.7} & 85.4  & 84.6 & 72.4 \\
    W/O \(LIF(\cdot)\)   & 80.0 & 86.9 & 88.2 & 88.8 & 89.4 & \textbf{90.0} \\\hline
    \textbf{Our work}   & 82.0 & 87.4 & 88.4 & 89.7 & 90.0 & \textbf{90.4}\\
    \hline
    \end{tabular}
\end{table}


\begin{figure*}[!t]
\centering
\subfloat[]{\includegraphics[width=0.45\linewidth]{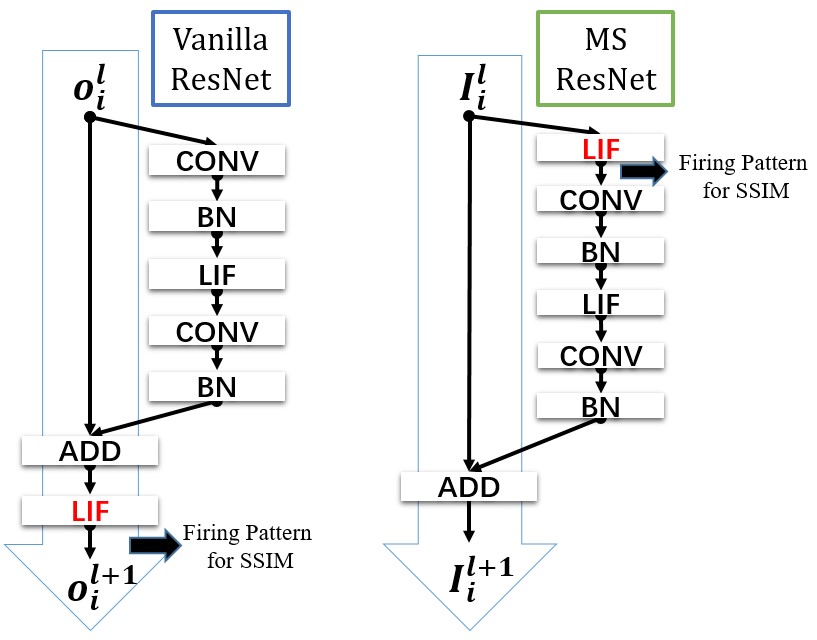}%
\label{fig:main_a}}
\hfil
\subfloat[]{\includegraphics[width=0.45\linewidth]{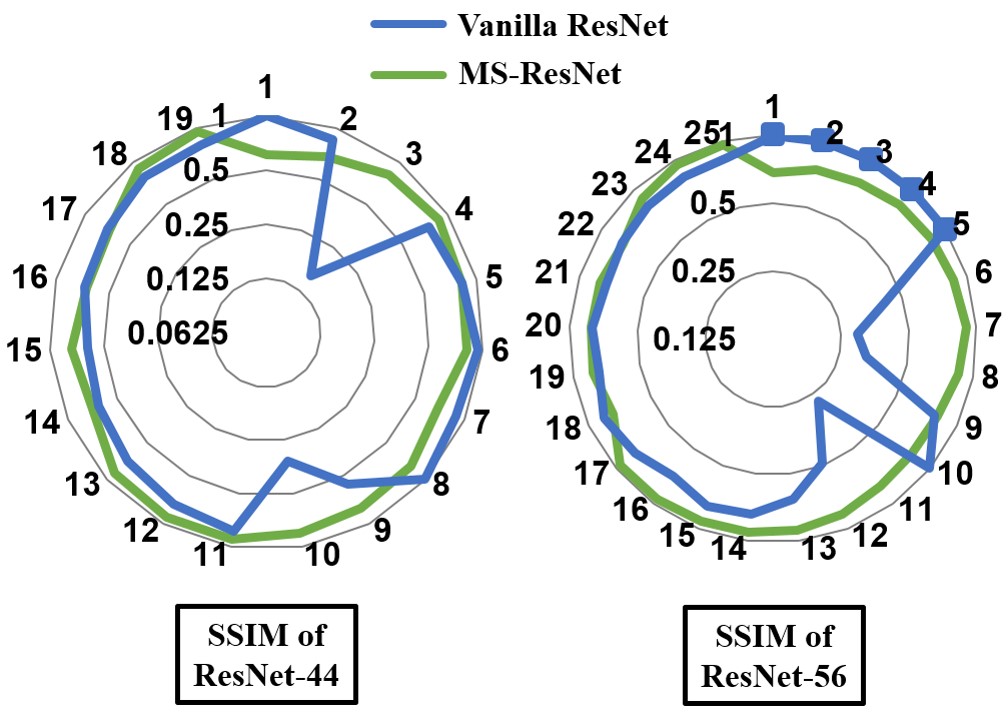}%
\label{fig:main_b}}
\caption{The vanilla spiking ResNet and proposed MS-ResNet. (a) The basic blocks of the two structures, where the arrows show the locations from which the firing patterns are extracted. (b) The radar charts of SSIM for spiking ResNet-44 and ResNet-56, respectively.}
\label{fig:Main}
\end{figure*}


\section{Spiking Residual Blocks}
\label{sec:Spiking Residual Blocks}
In this section, we will try to explain why the interblock \(\text{LIF}(\cdot)\) obstructs the applicability of vanilla ResNet in the specifics of SNNs' characteristics, and introduce the corresponding superiority of our model as well as some other concerns about the new design. The advantages of our model are threefold: 
\begin{itemize}
    \item An unimpeded inference flow which can avoid unavailing residual expression and workload imbalance.
    \item Achieving block dynamical isometry which is an important metric to avoid the gradient vanishing or explosion problem.
    \item Primary energy-efficient features which are intentionally maintained and will suit for further implementation on neuromorphic hardware.
\end{itemize}

The spiking residual block we proposed is illustrated in Figure~\ref{fig:main_a}. The interlayer \(\text{LIF}(\cdot)\) is removed to construct a shortcut that goes throughout the whole network and mainly deals with the confluences of residual paths from different blocks. Meanwhile, an additional \(\text{LIF}(\cdot)\) is placed at the top of the residual path to convert messages into sparse spikes and send them to subsequent neurons. The flow in the shortcut is conceptually closer to the sum of synaptic inputs to neuronal membranes, rather than spiking activity in the original structure, so we name the new structure as the Membrane-Shortcut ResNet (MS-ResNet) to put emphasize on the change in the shortcut.

\subsection{Residual Representation and Workload Balance at Inference}

Generally, in residual learning we would like a residual path to learn a relatively small perturbation with reference to the identity mapping of the shortcut and these perturbations will accumulate into an optimal transformation as the network deepens.

In the vanilla spiking ResNet, an identity mapping can be easily achieved by the block, i.e., a spike in the shortcut can always induce firing at the next layer and an inactive neuron will not activate its subsequent one, when we set the threshold $V_{th}$ smaller than 1 (e.g., 0.5) and the residual part is close to zero: 
\begin{equation}
    \boldsymbol{o_i^{l+1}}=\text{LIF}\left(\mathcal{F}\left(\boldsymbol{o^{l};\boldsymbol{W^{l+1}_i}}\right)+\boldsymbol{o_i^{l}}\right)\approx g\left(\boldsymbol{o_i^{l}}\right)=\boldsymbol{o_i^{l}}.
\end{equation}
Therefore, it can be inferred that to achieve identity mapping is not what really matters in the degradation problem of vanilla spiking ResNet. In regard of the accumulation by layers, we mainly focus on the changes in the output of each residual block and tend to examine what conditions need to be met when the firing state of a subsequent neuron will differ from the neuron imposing the identity mapping after receiving information from the residual path, i.e., $\left(o_i^{t,l},o_i^{t,l+1}\right) \in S = \left \{ s_i=\left.\left(o_i^{t,l},o_i^{t,l+1}\right) \right| o_i^{t,l}+o_i^{t,l+1}=1, o_i^{t}\in\left\{0,1\right\} \right \}$.

The probability of a successful change to the firing state can be decomposed into two conditions, which can be written as 
\begin{multline}
 \text{P}\left(s_i\in S\right) =\text{P}\left(\mathcal{F}(\boldsymbol{o^{t,l}})+o_i^{t,l}>V_{th}|o_i^{t,l}=0\right) \text{P}\left(o_i^{t,l}=0\right) \\
    +\text{P}\big(\mathcal{F}(\boldsymbol{o^{t,l}})+o_i^{t,l}<V_{th}|o_i^{t,l}=1\big)\text{P}\left(o_i^{t,l}=1\right),    
\end{multline}
where the part of decayed membrane potential is omitted here for simplicity. Assuming that the output of the residual path is a continuous random variable with a probability density function $\phi(u)$ and can be approximated as being independent of the input from a single neuron,  we have
\begin{multline}
    \text{P}\left(s_i \in S \right)= \text{P}\left(o_i^{t,l}=0\right) \int_{V_{th}}^{+\infty}\phi(u)du \\
    + \text{P}\left(o_i^{t,l}=1\right)\int_{-\infty}^{V_{th}-1} \phi(u) du.
\end{multline}
If the residual output satisfies a normal distribution $\mathcal{F}(\boldsymbol{o^{t,l}})\sim \mathcal{N}(0,\sigma_x^2)$, with $\sigma_x=V_{th}=0.5$  as adopted in Zheng et al.~\cite{Zheng_Wu_Deng_Hu_Li_2021}, we have $\text{P}\left(s_i\in S\right) \approx 16\% $. For the remaining 84\% of neurons, which occupy a significant portion, what the residual path conveys will not influence the current state at all and only contribute as decayed membrane potentials to later timesteps. Furthermore, the increment contributed by the residual path will be forgotten by those firing neurons due to the reset mechanism, making the residual representation of this block totally futile in both the spatial and temporal dimensions.

In particular, when considering only inputs from the neurons that do not undergo a state change between layer $l$ and layer $l+1$, we have
\begin{align}
o_i^{t,l+2}=&\text{LIF}\left(\mathcal{F} (\boldsymbol{o^{t,l+1};W^{l+2}})+o_i^{t,l+1}\right) \nonumber \\
    =&\text{LIF}\left(\mathcal{F}   (\boldsymbol{o^{t,l};W^{l+2}})+o_i^{t,l}\right),
\end{align}
which indicates that the residual path of the ${l\text{+1}}$-th layer does not contribute to the overall capacity of the network at timestep $t$ and we consider it as an unavailing residual representation. 

A straightforward solution could be increasing the variance of the residual output $\sigma_x^2$, which leads to higher probability of changing neuronal states and fewer unavailing residual paths. However, it is not our intention to discuss the optimal variance here, because this solution can never completely prevent unavailing residual representations and an overemphasis on the residual path is equivalent to decreasing the importance of the identity mapping, which disobeys the perturbation learning rule of ResNet. Actually the gating function in the interblock $\text{LIF}(\cdot)$ makes the output of a block somehow only 
a selection between the two, rather than the main shortcut aided by the residual path as we would expect.

While for MS-ResNet, the removing of the interblock \(\text{LIF}(\cdot)\) provides a clean path for the information flow and the confluences of residual paths will not be judged by the gating function, so there will be no unavailing residual representation, regardless of whether the firing state of a neuron actually changes. Small residual representations can always be accumulated as
\begin{align}
    \boldsymbol{I^{t,l+2}}=& \boldsymbol{I^{t,l+1}}+\mathcal{F}\left(\text{LIF}(\boldsymbol{I^{t,l+1}});\boldsymbol{W^{l+2}}\right) \nonumber \\
        =& \boldsymbol{I^{t,1}}+\sum_{k=1}^{l+1}\mathcal{F}\left(\text{LIF}(\boldsymbol{I^{t,k}});\boldsymbol{W^{k+1}}\right),
\end{align}
where $\boldsymbol{I^{t,l}}$ represents the synaptic input vector of the $l$-th layer.

To give a more intuitive verification about how firing patterns may change along with residual blocks of different layers, we adopt structural similarity index measure (SSIM) to quantify the similarity, which can be formulated as 
\begin{equation}
    SSIM(x,y)=\frac{(2\mu_x\mu_y+C_1)(2\sigma_{xy}+C_2)}{(\mu_x^2+\mu_y^2+C_1)(\sigma_x^2+\sigma_y^2+C_2)},
\end{equation}
with $\mu_x,\sigma_x^2,\mu_y,\sigma_y^2$ the mean and variance of image x and y, $\sigma_{xy}$ the covariance of x and y, and $C_1,C_2$ two stabilization constants. The pixel-values of an image are the firing rates of neurons from a feature map and the sampling point is illustrated in Figure~\ref{fig:main_a}. The value of SSIM ranges from -1 and +1, and only equals +1 if the two images are identical.

If the firing pattern varies evenly between layers, it will show up as a circle in the radar chart of Figure~\ref{fig:main_b}. Although a strict circle does not apply to the whole neural network because of the heterogeneity caused by the downsampling layers, our MS-ResNet has shown a more circular curve than vanilla spiking ResNet. Especially, there are five layers with an SSIM value of +1 in vanilla ResNet-56, which implies that the firing patterns do not change when information flows through the residual blocks and that these layers fail to help with feature extraction. Consequently, a following layer needs to compensate for the inaction of preceding layers, which manifests itself in the ensuing dramatic information change at $6$-th layer. The unbalanced workload across the network, especially when training deep networks, certainly reflects the irrationality of the spiking ResNet and gets effectively mitigated in MS-ResNet.

\subsection{Gradient Evolvement at Backpropagation}
Dynamical isometry, the equilibration of singular values of the input-output Jacobian matrix, has been developed in recent years as a theoretical explanation of well-behaved neural networks. 
In this subsection, we analyze with block dynamical isometry framework~\cite{chen2020comprehensive} that MS-ResNet can achieve gradient norm equality while the vanilla spiking ResNet cannot. 

Without loss of generality, a neural network can be viewed as a serial of blocks:
\begin{equation}
    \boldmath f(x_0)=f^L_{\theta^L}\circ f^{L-1}_{\theta^{L-1}} \circ \cdots \circ f^1_{\theta^1}(x_0),
    \label{eqn:serial}
\end{equation}
where $\boldsymbol{\theta^i}$ is the parameter matrix of the $i$-th layer. For the sake of simplicity, we denote $\frac{\partial\boldsymbol{f^{j}}}{\partial\boldsymbol{f^{j-1}}}$ as $\boldsymbol{J_j}$. Let $\phi(\boldsymbol{J})$ be the expectation of $tr(\boldsymbol{J})$ and $\varphi(\boldsymbol{J})$ be $\phi(\boldsymbol{J^2})-\phi(\boldsymbol{J})^2$. 

\begin{definition}[Block Dynamical Isometry]
    Consider a neural network that can be represented as Equation~(\text{\ref{eqn:serial}}) and the $j$-th block’s Jacobian matrix is denoted as $\boldsymbol{J_j}$. If $ \forall$ j, $\phi(\boldsymbol{J_j J_j}^T)$ $\approx 1$ and $\varphi(\boldsymbol{J_j J_j}^T) \approx 0$, the network achieves block dynamical isometry~\cite{chen2020comprehensive}.
\label{def:block dynamical isometry}
\end{definition}

\begin{lemma}
Assuming that for each of $L$ sequential blocks in a neural network, we have $\phi(\boldsymbol{J_iJ_i}^T)=\omega+\tau\phi(\widetilde{\boldsymbol{J_i}}\widetilde{\boldsymbol{J_i}}^T)$ where $\boldsymbol{J_i}$ is its Jacobian matrix. Given $\lambda \in \mathbb{N} ^+ <L$, if $C_L^\lambda (1-\omega)^\lambda$ and $C_L^\lambda\tau^\lambda$ are small enough, the network would be as stable as a $\lambda$-layer network when both networks have $\forall i $, $\phi(\boldsymbol{J_iJ_i^T}) \approx 1$~\cite{chen2020comprehensive}.
\label{lemma:shallow network trick}
\end{lemma}
Based on Definition~\ref{def:block dynamical isometry} and Lemma~\ref{lemma:shallow network trick}, we can judge whether a network can achieve gradient norm equality or not in the specifics of residual SNNs.

\begin{proposition}
Assuming that both neural networks consisting of $L$ sequential blocks, the vanilla spiking ResNet does not achieve block dynamical isometry while MS-ResNet can be as stable as a $\lambda$-layer network which satisfies $\phi(\boldsymbol{J_iJ_i^T}) \approx 1$ and $\lambda\in \mathbb{N^+} <L$.
\end{proposition}
\begin{proof}
According to Lemma~\ref{lemma:multiplication} in Appendix, the Jacobian matrix of the whole network can be decomposed into the multiplication of its blocks' Jacobian matrices. We expect that each block should satisfy $\forall~i,\phi\left(\boldsymbol{J_iJ_i}^T\right) \approx 1$, so as to provide a stable gradient evolvement. The common components of neural networks and their spectrum-moment have been summarized in~\cite{chen2020comprehensive}, so here we mainly focus on the analysis of $\text{LIF}(\cdot)$. 

The gate function defined in Equation~(\ref{Eqn:gate_function}) and its surrogate gradient in Equation~(\ref{Eqn:Surrogate_func}) are both element-wise operation, so the Jacobian matrix $\boldsymbol{J_g}$ of $g(\cdot)$ is a diagonal matrix whose elements are either 0 or $\frac{1}{a}$. Its probability density function satisfies
\begin{equation}
    \rho_{\boldsymbol{J_g}} \left(z\right)=\left(1-p\right)\delta\left(z\right)+p\delta\left(z-\frac{1}{a}\right)
\end{equation}
where $p$ denotes the probability that $(x-V_{th})\leq \frac{a}{2}$ is true.
Thus we have
\begin{align}
        \phi\left(\boldsymbol{J_gJ_g}^T\right)&=\int_{\mathbb{R}}{z\left(\left(1-p\right)\delta\left(z\right)+p\delta\left(z-\frac{1}{a^2}\right)\right)dz}\nonumber \\
        &=\frac{p}{a^2},
\end{align}
\begin{align}
        \varphi\left(\boldsymbol{J_gJ_g}^T\right)&=\int_{\mathbb{R}} z^2\left(\left(1-p\right)\delta\left(z\right)+p\delta\left(z-\frac{1}{a^2}\right)\right)dz \nonumber \\ 
     &- \phi\left(\boldsymbol{J_gJ_g}^T\right)^2
    =\frac{p-p^2}{a^4}.
\end{align}
$a$ is a fixed global hyper-parameter, while the probability $p$ will change dynamically during the training process along with the weight updates and different input data, so it turns out to be difficult to match the two between different layers. 
Under such conditions, the gate function makes \(\text{LIF}(\cdot)\) not qualified as a serial block for Definition~\ref{def:block dynamical isometry}.

We denote the Jacobian matrix of the residual path in a block as $\widetilde{\boldsymbol{J_i}}$. When the interblock $\text{LIF}(\cdot)$ is not included, according to Lemma~\ref{lemma:addition} in Appendix we have $\phi(\boldsymbol{J_iJ_i}^T)=1+\gamma^2 \phi(\widetilde{\boldsymbol{J_i}}\widetilde{\boldsymbol{J_i}}^T)$, where $\gamma$ is from the linear scale  transformation $\gamma x+\beta$ within the normalization at the bottom of a residual path. MS-ResNet can be viewed as an extreme example of Lemma~1 with $(1-\omega) \rightarrow 0$. Therefore $\forall \lambda$, $C_L^\lambda(1-\omega)^\lambda$ is close to zero, and  $C_L^\lambda\gamma^\lambda$ can be small enough for a given $\lambda$ if $\gamma$ is initialized as a relative small value. In this way, the error of non-optimal block will be influential only within $\lambda$ layers and the MS-ResNet will be as stable as a much shallower $\lambda$-layer network.

While for a residual block of vanilla spiking ResNet, we have $\phi(\boldsymbol{J_iJ_i}^T)=\left(1+\gamma^2 \phi(\widetilde{\boldsymbol{J_i}}\widetilde{\boldsymbol{J_i}}^T)\right)\cdot \phi\left(\boldsymbol{J_gJ_g}^T\right)=\left(\frac{p}{a^2}+\gamma^2\frac{p}{a^2}\phi(\widetilde{\boldsymbol{J_i}}\widetilde{\boldsymbol{J_i}}^T)\right)$, which indicates the gradient flow of vanilla spiking ResNet is hard to maintain stable and may manifest itself as the degradation problem.
\end{proof}

In a nutshell, we find that \(\text{LIF}(\cdot)\) is not qualified as a serial function between blocks for achieving block dynamical isometry, which may manifest itself as the degradation problem. However, MS-ResNet can avoid this drawback and attain great stability constituting a much shallower network in effect than it appears to be for gradient norm.

\subsection{Spike-based Convolution and an Extra $\text{LIF}(\cdot)$ at the Top}
It should be pointed out that vanilla ResNet without interblock \(\text{LIF}(\cdot)\) is also undesired for SNNs. One main source of energy efficiency for neuromorphic computing is the spike-based convolution (CONV) which means that the CONV layers will receive and process binary spike inputs and it is feasible to replace multiply-and-accumulate (MAC) operations in ANNs with spike-driven synaptic accumulate (AC) operations in SNNs. Specialized optimization, for example the look up table~\cite{liang2021h2learn}, can be consequently applied to further boost its efficiency on neuromorphic devices. However once interblock \(\text{LIF}(\cdot)\) is removed, the CONV at the top of the next block will receive continuous inputs rather than binary spikes, causing difficulty in benefiting from spike-based operations and rich input/output sparsity.
 
 In addition, removing the interblock \(\text{LIF}(\cdot)\) will make the CONV layer in the residual path connected with the one at the top of its subsequent block, i.e. form a structure of \{CONV-BN-CONV-BN-LIF\} between two adjacent blocks. These two CONV3x3 are equal to a single CONV5x5 in effect, which we speculate will make the network actually a shallower structure and thus weaken its feature extraction ability. It turns out that the accuracy of W/O \(\text{LIF}(\cdot)\) will be lower than the vanilla model before the degradation problem occurs (Table~\ref{tab:depth_analysis}). Therefore, we place an extra \(\text{LIF}(\cdot)\) at the top of each residual path in order to maintain spike-based CONV as well as fully utilize the CONV layers.
 
Also note that an extra $\text{LIF} (\cdot)$ is placed at the end of the entire convolutional part, to ensure that even if a portion of the information flows only through the shortcut connections, the subsequent
FC classifier will still receive spiking signals.
 
\subsection{Depth Analysis on CIFAR-10}
We carry out the depth analysis experiment on CIFAR-10 with MS-ResNet as well. The results are shown in Table~\ref{tab:depth_analysis}, which indicate that our MS-ResNet can expand to a larger scale without facing the degradation problem of vanilla spiking ResNet, while maintaining better accuracy of the shallow networks compared to W/O $\text{LIF}(\cdot)$. Besides, to avoid the possibility that MS-ResNet can only shift its accuracy peak to a certain extent like vanilla spiking ResNet, we set $n$ to 36 and 80 to obtain extremely deep MS-ResNet110 and MS-ResNet482 on CIFAR-10, with their test accuracy of 91.7\% and 91.9\% respectively. Although the improvement in test accuracy is not significant due to the limited  regularization method and the resulting overfitting problem, it surely evidences the scalability and the capability of our model to avoid degradation.

\section{Experiments}
\label{sec:Experiments}
\subsection{Benchmark Results}
\begin{table*}[!t]
\caption{Results on ImageNet.}
\label{tab:Image_results}
\centering
\begin{tabular}{clccc}
\hline
Method                             & Work                                            & Model           & timestep              & Acc.(\%) \\ \hline
\multirow{7}{*}{\textbf{\begin{tabular}[c]{@{}c@{}}ANN\\ Conversion\end{tabular}}}  & Sengupta et al.~\cite{Senguputa_2013_going_deeper}       & ResNet-34       & 2500                      & 65.47    \\ 
    & Han, Srinivasan, and Roy~\cite{Han_2020_CVPR}          & ResNet-34    & 4096                                    & 69.89    \\
                                   & \multirow{2}{*}{Li et al.~\cite{Li_freelunch_2021}}                  & \multirow{2}{*}{ResNet-34}       &  256 & 74.61    \\
                                   &                                                 &        & 32                      & 64.54    \\
                                   
                                   & \multirow{2}{*}{Hu et al.~\cite{hu2018spiking}}                  & ResNet-34       & 350 & 71.61    \\
                                   &                                                 & ResNet-50       &        350               & 72.75    \\ 
                                   & St{\"o}ckl and Maass~\cite{stockl2021optimized}\dag            & ResNet-50       & 500                   & 75.10    \\
                                   \hline
\multirow{4}{*}{\textbf{\begin{tabular}[c]{@{}c@{}}Direct \\ Training\end{tabular}}}   & \multirow{2}{*}{Zheng et al.~\cite{Zheng_Wu_Deng_Hu_Li_2021}}  & ResNet-50       & \multirow{2}{*}{6}    & 64.88    \\
                                   &                                                 & Wide-ResNet-34  &                       & 67.05    \\ 
                                   & \multirow{2}{*}{Fang et al.~\cite{fang2021deep}}                   & ResNet-34       & \multirow{2}{*}{4}    & 67.04    \\
                                   &                                                 & ResNet-101      &                       & 68.76    \\ \hline
\multirow{4}{*}{\textbf{\begin{tabular}[c]{@{}c@{}}Direct\\  Training \end{tabular}}}     & \multirow{4}{*}{\textbf{\begin{tabular}[c]{@{}c@{}} Our Work\\   \boldmath$\text{\ MS-ResNet}$ \unboldmath \end{tabular}}}                       & ResNet-18       & \multirow{2}{*}{6}    & 63.10    \\
                                   &                                                 & ResNet-34       &                       & 69.42    \\
                                   &                                                 & ResNet-104      & \multirow{2}{*}{5}     & \textbf{74.21}    \\
                                   &                                                 & ResNet-104*      &                       & \textbf{76.02}    \\\hline
\multirow{3}{*}{\textbf{Backpropagation}}                & \multirow{3}{*}{ANN}                            & ResNet-18       & \multirow{3}{*}{/}    & 69.76    \\
                                   &                                                 & ResNet-34       &                       & 73.30     \\
                                   &                                                 & ResNet-104\ddag      &                       & \textbf{76.87}    \\ 
                                   \hline
\multicolumn{5}{l}{\footnotesize \dag A spike is allowed to carry multi-bit information.*The input crops are enlarged to 288$\times$288 in inference.}\\
\multicolumn{5}{l}{\ddag Since ResNet-104 is not a standard ResNet model, we train its ANN counterpart under the same recipe.}
\end{tabular}
\end{table*}

\subsubsection{ImageNet}

We evaluate our models on the ImageNet 2012 dataset~\cite{imagenet}. The models are trained with 1.28M training images and tested with 50K validation images. The training recipe at first simply follows that of He et al.~\cite{He_2016_resnet} for $\text{MS-ResNet18}$ and $\text{MS-ResNet34}$. However, we find MS-ResNet104 would suffer from severe overfitting. Despite little accuracy improvement on the validation set, its training accuracy improves by a significantly larger margin. We hypothesize that it is because MS-ResNet104 has much higher model capacity. To fully unleash the potential of the deep spiking model $\text{MS-ResNet104}$, an advanced training recipe is taken with stronger data augmentation and regularization, and a two-phase training process consisting of a T=1 pretrain phase and a formal training phase is proposed to save training time (See \textbf{Appendix~\ref{appendix_exp}} for more training details). 

 A satisfactory upward trend in accuracy has been observed as the network deepens in Table~\ref{tab:Image_results}. It evidences the scalability of our models and clearly demonstrates the potential of deepening SNN models. When compared with other advanced works, MS-ResNet34 with an accuracy of \textbf{69.42\%} surpasses all previous directly-trained SNNs with the same depth, and MS-ResNet104 with an accuracy of \textbf{74.21\%} is even comparable to the converted SNNs in spite of much fewer timesteps required for inference. To our best knowledge, this is the first time such high performance is reported on ImageNet with directly-trained SNNs. We also find that just enlarging the images for inference from 224$\times$224 to 288$\times$288 can improve the accuracy to a more competitive score at 76.02\%.
\begin{table}[!t]
    \caption{Results on CIFAR10-DVS.}
    \label{tab:CIFAR10-DVS}
    \centering
    \begin{tabular}{lrrr}
    \hline
    Work       & Method                                       & Params     & Acc.(\%)       \\ \hline
    Ramesh et al.~\cite{Ramesh_2019_DART}          & DART                    &N.A.                      & 65.78          \\
    Kugele et al.~\cite{kugele2020efficient}          & Rollout-ANN                 &0.5M                       & 66.75          \\
    Zheng et al.~\cite{Zheng_Wu_Deng_Hu_Li_2021}          & Spiking ResNet-19     &12.6M                        & 67.80          \\
    Yao et al.~\cite{yao2021temporal}          & TA-SNN                          &1.7M              & 72.00          \\
    Fang et al.~\cite{fang2020incorporating}          & PLIF                    &1.5M                      & 74.80          \\ \hline
    \textbf{Our work} & \textbf{MS-ResNet20} & 0.27M   & \textbf{75.56} \\ \hline
    \end{tabular}
\end{table}

\subsubsection{CIFAR10-DVS}
CIFAR10-DVS is an event-stream dataset for object classification~\cite{CIFAR10-DVS}. 10,000 frame-based images from CIFAR-10 are recorded by a dynamic vision sensor (DVS) and converted into event streams. We take the dataset to test our models for their capability of processing spatiotemporal event data and adopt the same data pre-processing in~\cite{fang2020incorporating}.

Our MS-ResNet20 achieves a new record on CIFAR10-DVS (Table~\ref{tab:CIFAR10-DVS}). The model mainly follows the paradigm in Table~\ref{tab:narrownet} except that an additional downsampling is placed at the first CONV stage due to the larger input size. Our model, despite its depth, actually has a smaller number of parameters, which is about only one-sixth of those in Yao et al.~\cite{yao2021temporal} and Fang et al.~\cite{fang2020incorporating}.

\begin{table}[]
    \centering
    \caption{Energy Cost for Processing a Single Sample of ImageNet and CIFAR10-DVS.}
    \label{tab:enegy_estimate}
    \begin{tabular}{lrrrrr}
\hline
\multicolumn{6}{c}{\textbf{32bit-FP~\cite{Horowitz_2014_energy}:\quad MAC\ 4.6pJ \quad AC \ 0.9pJ}}                                                                        \\ \hline
\multicolumn{6}{c}{ImageNet}                                                                                 \\ \hline
Model      & FLOPs  & E(ANN)                         & SyOPs  & E(SNN)                       & $\frac{\text{E(ANN)}}{\text{E(SNN)}}$ \\ \hline
ResNet-34  & 3.53G  & 16.22mJ                        & 4.77G  & 4.29mJ                       & 3.78          \\
ResNet-104 & 11.79G & 54.24mJ                        & 11.32G & 10.19mJ                      & 5.32          \\ \hline
\multicolumn{6}{c}{CIFAR10-DVS}                                                                             \\ \hline
ResNet-20  & 40.11M & 184.46\textmu J & 8.32M  & 7.46\textmu J & 24.73         \\ \hline
\end{tabular}
\end{table}

\subsection{Firing Patterns of MS-ResNet}
\begin{figure}[!t]
\centering
\subfloat[]{\includegraphics[width=\linewidth]{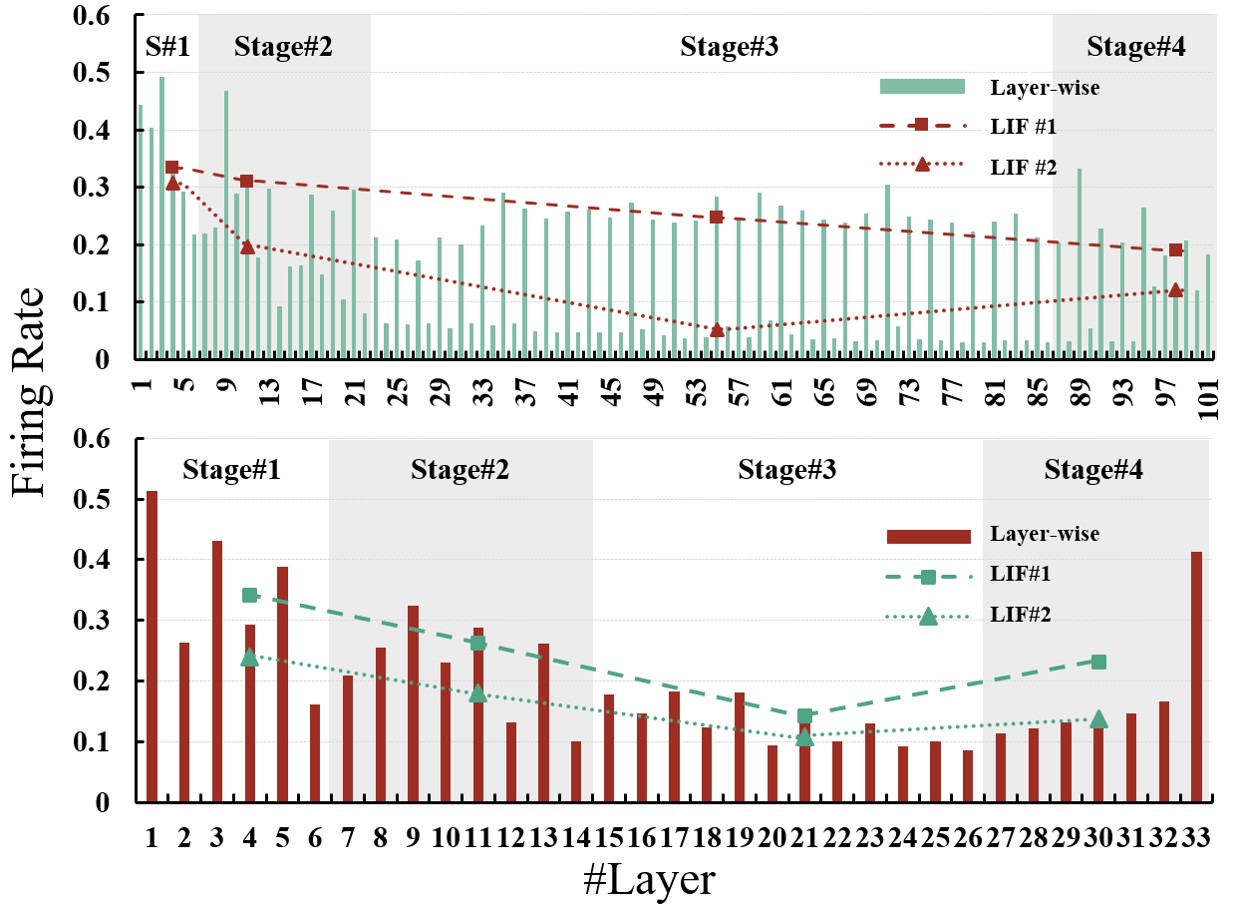}%
\label{fig:Firing_rate}}
\hfil
\subfloat[]{\includegraphics[width=\linewidth]{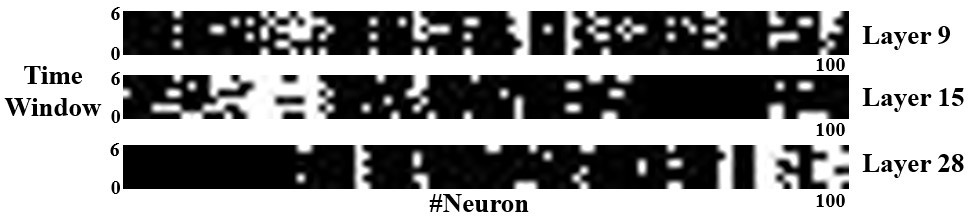}%
\label{fig:raster}}
\caption{Firing Patterns. (a) The layer-wise firing rates in MS-ResNet34 and MS-ResNet104. The triangular and square data points represent the firing rates of two LIFs averaged in each stage, respectively. The stages are divided according to the sizes of feature maps. (b) The firing patterns of 100 neurons from 3 layers of MS-ResNet34, where the white dots represent firing.}
\label{fig_sim}
\end{figure}

High sparsity is observed in MS-ResNet. The firing rate is defined as the firing probability of each neuron per timestep, and is estimated under a batch of random samples by $r = \frac{\#\text{Spike}}{\# \text{Neuron} \cdot T}$, where \#Spike denotes the number of spikes during $T$ timesteps and \#Neuron  denotes the number of neurons in the network. The firing rates of MS-ResNet34 and MS-ResNet104 are \textbf{0.225} and \textbf{0.192}, respectively. Given the five to six timesteps required, MS-ResNet maintains such sparsity that only about one spike per neuron is emitted on average during the full processing of a single ImageNet sample. 

 Figure~\ref{fig:Firing_rate} shows layer-by-layer firing rates. On the whole, the firing rate remains relatively stable within each stage, gradually decreases in the first three stages, and possibly picks up in the last stage. Interestingly, it can be noticed that a certain pattern exists in the alternate layers that neurons in the second $\text{LIF}(\cdot)$ of each residual path tend to be less active than those in the first one. In particular, in the third stage of MS-ResNet104, the average firing rate of the second $\text{LIF}(\cdot)$ can be as low as 5.2\%, while for the first $\text{LIF}(\cdot)$ it is almost five times higher, which indicates that our residual block has a relatively active representation in the shortcut (if we convert the flow of synaptic inputs into spikes), while the spiking activity in the residual branch is kept calm and sparse. A more detailed raster diagram of spikes is shown in Figure~\ref{fig:raster}. It shows that the firing of a neuron can be regular and the low firing rate is mainly affected by the large number of silent neurons, which can fit well with the event-driven neuromorphic chips.

\subsection{Energy Efficiency Estimation}
As mentioned earlier, one of the features intentionally maintained in our model is the spike-based CONV. To further demonstrate the energy efficiency, we estimate the energy cost based on the number of operations and the data for various operations in 45nm technology~\cite{Horowitz_2014_energy}. Our main focus is on the CONV layers in residual paths, which constitute a major part of floating-point operations (FLOPs) of the network. The estimation does not include the encoding layer that processes images and converts them into spike trains, and the downsampling layers in the shortcuts, because they do not meet the spike-based CONV requirement. Nevertheless, their fixed cost only accounts for 4\% of FLOPs in ResNet-34 and 1\% in ResNet-104. 

 As the network deepens, the ratio of energy consumption $\frac{E(SNN)}{E(ANN)}$ for a single image will approach a value of $T*firing\_rate*\frac{E(AC)}{E(MAC)}$. With the sparsity of firing and the short simulation period, MS-ResNet can achieve the calculation with about the same number of synaptic operations (SyOPs) rather than FLOPs (Table~\ref{tab:enegy_estimate}), which corresponds to that each neuron emits only one spike on average. For a single ImageNet image, the energy cost in an SNN are a third to a fifth of that in the ANN with the same structure.

When dealing with spatiotemporal stream data, ANNs are expected to possess an ability of sequence processing, and consequently recurrent connections are introduced for memorizing previous states and behaving temporal dynamics~\cite{he2020comparing}. Different from the process of a static image, both recurrent ANNs and SNNs will spend multiple timesteps to perform frame-based information integral for tasks of sequence learning. Notably the number of timesteps $T$ will not contribute as a factor to the energy consumption ratio $\frac{E(SNN)}{E(ANN)}$, so the energy efficiency will be more prominent. The average firing rate of MS-ResNet20 is 22.19\% for CIFAR10-DVS. With the binary and sparse spiking activity, the energy consumption of processing each frame of CIFAR10-DVS in SNNs is one twenty-fourth of that in ANNs.

\section{Discussion}
\label{sec:Discussion}
\subsection{The Loss Landscape of MS-ResNet}
\begin{figure*}[htbp]
    \centering
    \includegraphics[width=0.8\textwidth]{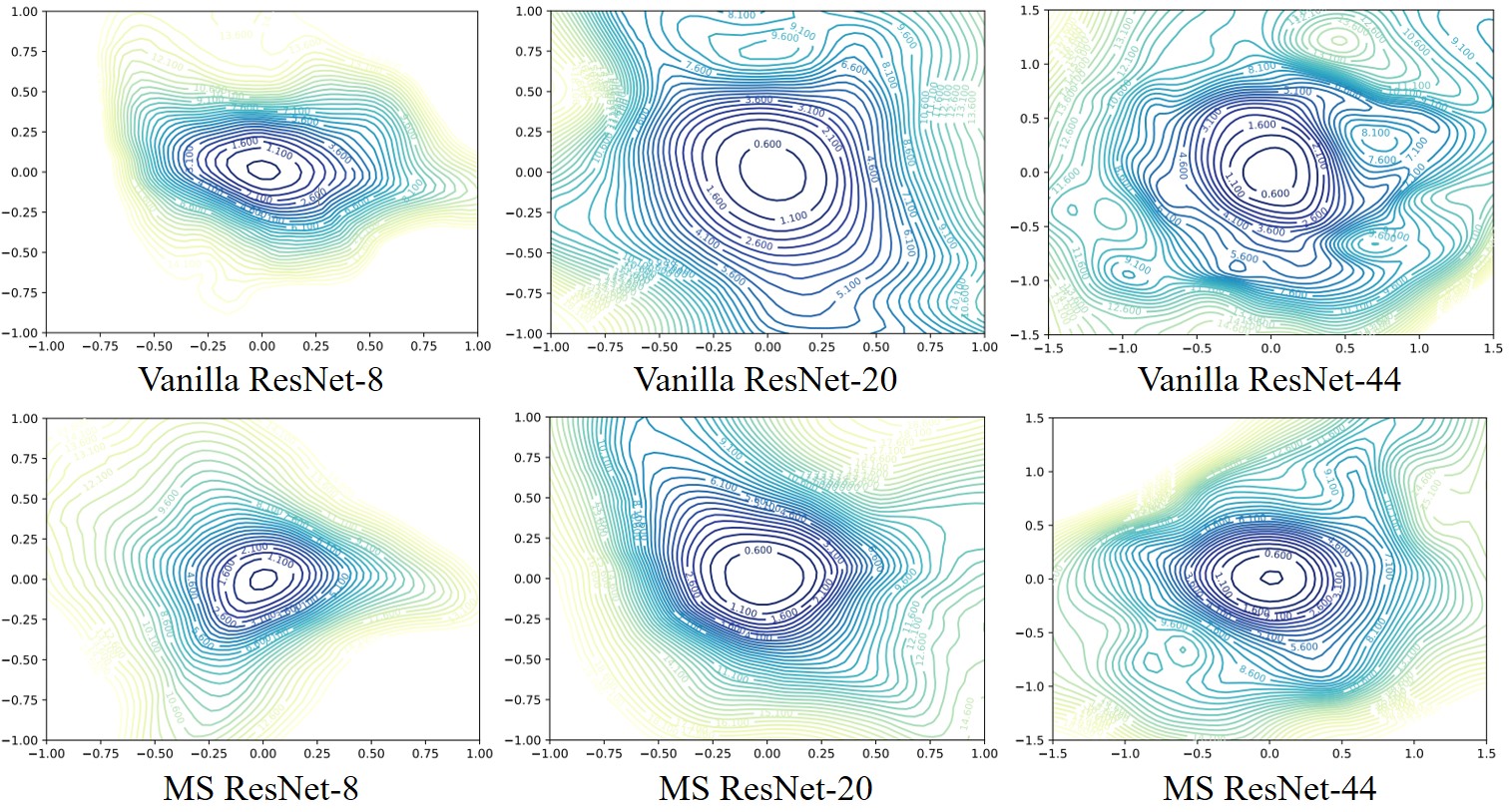}
    \caption{2D visualization of the loss surfaces of vanilla spiking ResNet and MS-ResNet with different depths. Note that for a 44-layer network, the area is wider and the coordinates are extended to provide a better view of the landscape.}
    \label{fig:loss_landscape}
\end{figure*}

We further visualize the local loss landscapes around the minima to which different spiking neural networks eventually converge, in order to explore how structural modification might affect the training of networks. We plot a loss function of the form $f(\alpha,\beta)=L(\theta^*+\alpha \delta+\beta \eta)$, with $\theta^*$ the set of trained parameters and $\delta,\eta$ two orthogonal direction vectors which are filter-wise normalized as in \cite{li2018visualizing}. 

From Figure~\ref{fig:loss_landscape} we see that the loss landscape of a shallow network is typically dominated by a region with convex contours in the center, and has no significant non-convexity, which is the same in both vanilla spiking ResNet and MS-ResNet. As the network gets deeper, the basin area that represents relatively well-behaved networks gradually expands, which we believe reflects the potential of a deep model for a possible superior local minimum and better generalization ability. However, as the vanilla spiking ResNet goes deeper, the appearance of more local minima and maxima makes the non-convexity stronger, and the direction of the gradient (which is perpendicular to the contour line in the figure) also deviates significantly from the central minimum, resulting in poor training and test accuracy. In MS-ResNet, the structural modification prevents the model from falling into ill-conditioned loss cases, thus guaranteeing the successful training of deep models and avoiding the degradation problem. This observation can be seen as supporting evidence for our reasoning in Section IV.

\subsection{The Initialization of the Residual Path}

\begin{figure}[!t]
\centering
\subfloat[]{\includegraphics[width=0.5\linewidth]{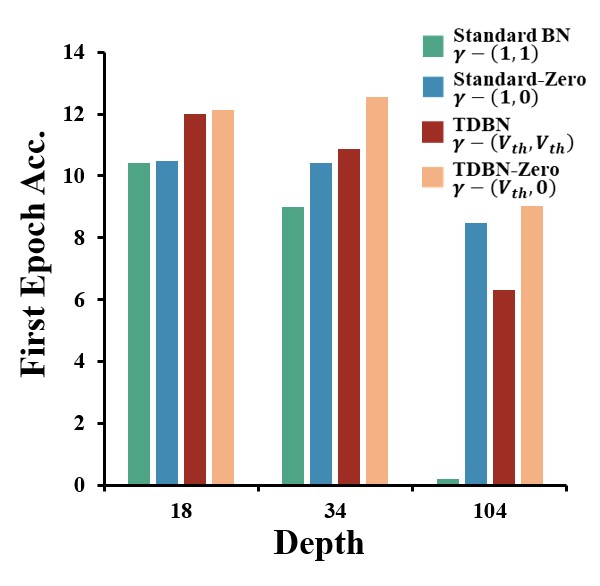}%
\label{fig:Init_a}}
\hfil
\subfloat[]{\includegraphics[width=0.45\linewidth]{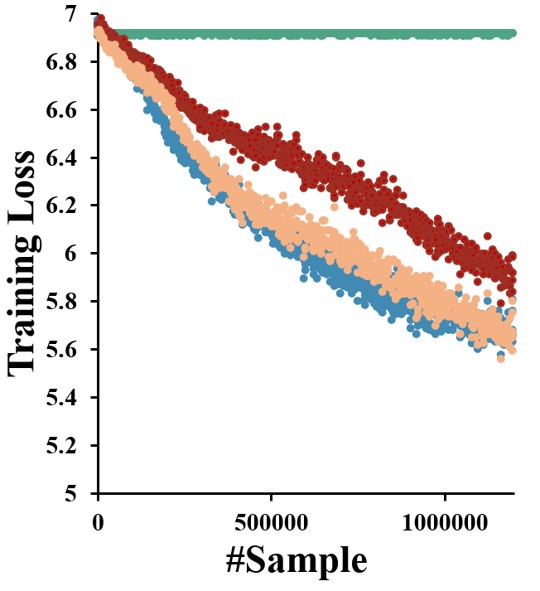}%
\label{fig:Init_b}}
\caption{(a) The accuracy results of models after the first epoch of training under different initialization, where the gammas are  in the form of $\left(\gamma_1,\gamma_2\right)$. (b) The curves of training loss with 104 layers, where the loss of MS-ResNet104 actually does not show a decrease in even 10 epochs when it is initialized by standard BN.}
\label{fig_Init}
\end{figure}
Weight initialization can be an important consideration in the design of a neural network model. In spiking neural networks, the distribution of the synaptic input is expected to be modulated by weight initialization to provide a stable firing pattern across layers, and it will be crucial to the early convergence speed of the network and the stability of the training process. In this subsection, we introduce a simple and effective initialization trick based on batchnorm suitable for the training of deep MS-ResNet.

The BN layer performs a linear scale transformation after standardizing its input $\gamma \widehat x+\beta$. For simplicity, $\gamma_1, \gamma_2$ are used to denote the linear parameter of the first and second BN in each residual block, respectively. In the proof of \textbf{Proposition 1}, we have mentioned that $\gamma_2$ can help to achieve block dynamical isometry in MS-ResNet when it is set close to zero. Here we compare the training curves and accuracies of MS-ResNet at the first epoch under different initialization conditions in Figure~\ref{fig_Init}. It can be seen that assigning $\gamma_2$ to zero exhibits significant acceleration of convergence, both for the standard BN and TDBN. Especially when the convergence becomes more difficult as the network deepens to 104 layers, it allows one non-convergent case to be radically improved. We speculate that initializing $\gamma_2$ to zero will give the model a clean and branch-less starting point and thus provide a faster convergence. This trick has also been observed in ANNs~\cite{He_2019_bag_of_tricks}.

It should be noted that it is inappropriate to initialize $\gamma_1$ to a small value close to zero. We have observed in our experiments that this will have the opposite effect to the method above. The main reason is that this initialization causes the output of the first LIF layer to all be zeros, while the zero-variance synaptic input prevents the second BN from working properly and the normalization process would be equivalent to multiplying by an extremely large $\gamma_2$.

\subsection{Alternative Structures}
\begin{figure}
    \centering
    \includegraphics[width=0.9\linewidth]{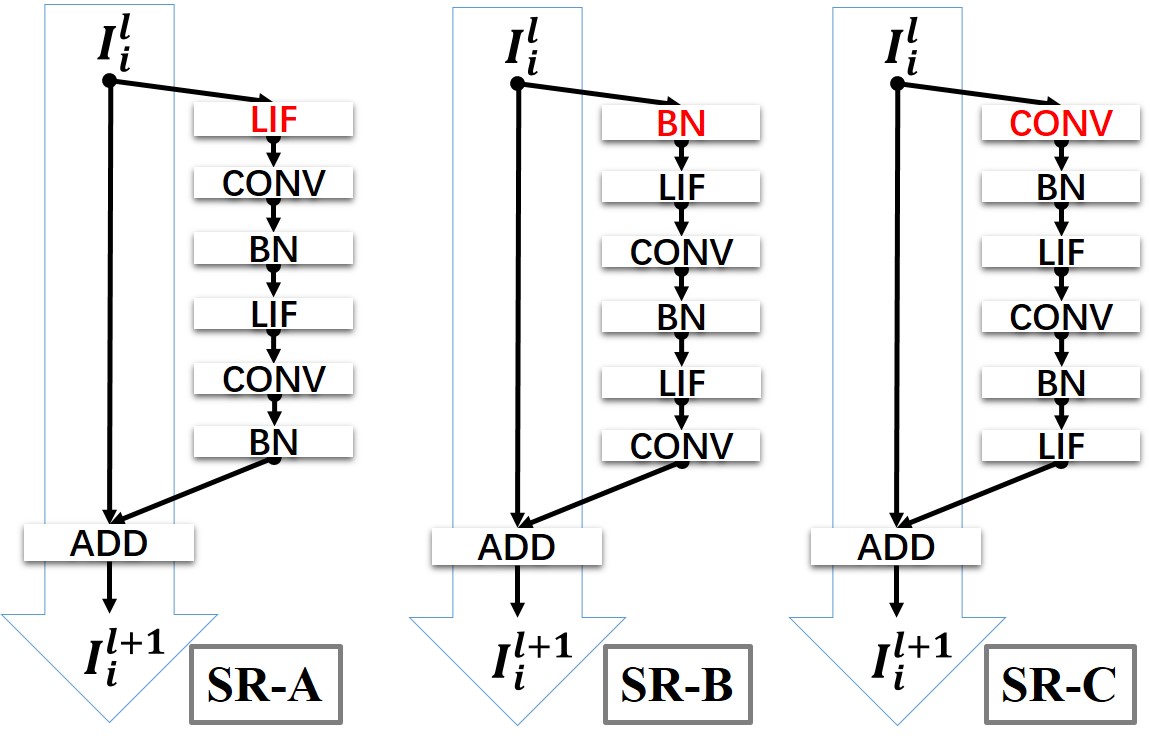}    \caption{Alternative structures with different shortcut-insertion positions, where SR stands for Spiking ResNet and SR-A corresponds to MS-ResNet adopted by us. }
    \label{fig:alternatives}
\end{figure}
There exist several alternative structures based on the removal of interblock $\text{LIF}(\cdot)$. In this subsection, we provide a brief discussion on the alternative structures obtained by inserting shortcuts at different positions as illustrated in Figure~\ref{fig:alternatives}.

From the results of depth analysis in Table~\ref{tab:alternatives}, despite the slight differences in the residual paths, the three structures can all expand to a large scale without facing the degradation problem. This emphasizes the importance of the removal of interblock $\text{LIF}(\cdot)$ and keep a thorough identity mapping in the whole network, as observed in deep ANN structures~\cite{He_2016_identitymapping}. Nevertheless, it should be noted that rather than increasing accuracy by a slight margin on ANNs, adopting such a design results in significant improvements in both network scalability and task performance of SNNs.

With regard to the selection between the three, we mainly take computing efficiency into our consideration. One obvious drawback of SR-C is that it does not meet the spike-based CONV criterion. SR-B is close to SR-A as a good choice. But the accuracy of SR-B is a little behind and the separation of CONV and BN in SR-B causes difficulty in utilizing the BN fusion technique for further implementation. Therefore, we take SR-A as our final choice and its effectiveness has been verified through experiments in Section V.

\begin{table}[!t]
    \centering
    \caption{Depth Analysis of Alternative Structures.}
    \label{tab:alternatives}
\begin{tabular}{ccccc}
\hline
Dataset                    & Depth & SR-A  & SR-B  & SR-C  \\\hline
\multirow{4}{*}{CIFAR-10}  & 32    & 90.59 & 90.04 & 89.48 \\
                           & 44    & 90.96 & 90.82 & 90.67 \\
                           & 56    & 91.33 & 91.18 & 90.78 \\
                           & 110   & 91.72 & 91.65 & 92.12 \\\hline
\multirow{4}{*}{CIFAR-100} & 32    & 61.35 & 61.06 & 61.03 \\
                           & 44    & 63.84 & 62.34 & 62.35 \\
                           & 56    & 65.24 & 63.94 & 63.73 \\
                           & 110   & 66.83 & 66.03 & 66.82\\ \hline
\end{tabular}
\end{table}

\section{Conclusion}
\label{sec:Conclusion}
In this work, we report the degradation problem when applying vanilla ResNet and reveal the implicit unavailing residual representation and unstable gradient norm caused by the interblock $\text{LIF}(\cdot)$. Accordingly, we propose a novel spiking residual block, MS-ResNet, which effectively mitigates the degradation problem and enables the direct training of a 482-layer model on CIFAR-10 and a 104-layer model on ImageNet. The great depth brings superior representation power. To our best knowledge, this is the first time such high performance is reported on ImageNet with directly-trained SNNs. In addition, our resulting models attain very sparse spiking activity and extremely short simulation, indicating remarkable energy efficiency especially for processing spatiotemporal information. We believe a deep and powerful SNN model is surely to work as the backbone for downstream tasks such as object detection, semantic segmentation and event-based optical flow estimation, and facilitate the exploration of brain-inspired computing in future.


\appendices

\section{Gradient Evolvement at Backpropagation}
Under the assumption that the Jacobian matrices of different blocks are independent, for a general serial neural network the following two lemmas hold:
\begin{lemma}[Multiplication](Theorem 4.1 in~\cite{chen2020comprehensive})
Given $\boldsymbol{J}:=\prod_{i=L}^1 \boldsymbol{J_i}$, where $\{\boldsymbol{J_i}\in \mathbb{R}^{m_i\times m_{i-1}}\}$ is a series of independent random matrices. If $(\prod_{i=L}^1\boldsymbol{J_i})(\prod_{i=L}^1\boldsymbol{J_i})^T$ is at least the $1^{st}$ moment unitarily invariant, we have 
\begin{equation}
    \phi\left((\prod_{i=L}^1\boldsymbol{J_i})(\prod_{i=L}^1 \boldsymbol{J_i})^T \right)=\prod_{i=L}^1\phi(\boldsymbol{J_i J_i}^T).
\end{equation}
\label{lemma:multiplication}
\end{lemma}

\begin{lemma}[Addition] (Theorem 4.2 in \cite{chen2020comprehensive}) Given $\boldsymbol{J}:=\prod_{i=L}^1 \boldsymbol{J_i}$, where $\{\boldsymbol{J_i}\in \mathbb{R}^{m_i\times m_{i-1}}\}$ is a series of independent random matrices. If at most one matrix in $\boldsymbol{J_i}$ is not a central matrix, we have
\begin{equation}
    \phi(\boldsymbol{JJ}^T)=\sum_i\phi(\boldsymbol{J_iJ_i}^T).
\end{equation}
\label{lemma:addition}
\end{lemma}

\section{Experimental Details}
\label{appendix_exp}
The source code in Pytorch is available at \href{https://github.com/Ariande1/MS-ResNet.}{https://github.com/Ariande1/MS-ResNet.}

\begin{table}[htbp]
    \centering
    \caption{Neuronal Configuration Parameters of LIF.}
    \label{tab:config}
    \begin{tabular}{cc}
    \hline
    \multicolumn{1}{l}{Parameter} & \multicolumn{1}{l}{Value} \\ \hline
    $V_{th}$                           & 0.5                       \\
    $V_{reset}$                      & 0                         \\
    $\tau_{mem}$                           & 0.25                      \\
        $T$                           & 6                      \\
    $a$                             & 1                         \\ \hline
    \end{tabular}
\end{table}
\subsection{CIFAR-10 Training Details}
\label{appendix_cifar10}

Configuration parameters about the LIF model are shown in Table~\ref{tab:config}, and  kept in all experiments of this work.

By setting $n$ to $5,7,9,18,80$ in Table~\ref{tab:narrownet}, we obtain our spiking ResNet-32,44,56,110,482. The hyper-parameter settings for depth analysis of Table~\ref{tab:depth_analysis} and \ref{tab:alternatives} are detailed in Table~\ref{tab:setting_CIFAR10}. The data pre-processing and weight decay are strengthened in the depth analysis of alternative structures due to great depths of models.
The experiment is conducted on NVIDIA RTX 2080Ti GPUs and synchronized BN technique will be used if multiple GPUs are required. We report the average test accuracy of the last five epochs as our results.
\begin{table}[!t]
    \centering
    \caption{Hyper-parameters for Experiments on CIFAR-10.}
    \label{tab:setting_CIFAR10}
\begin{tabular}{cccc}
\hline
Experiment      & Component                                                                    & Name        & Value \\ \hline
\multirow{9}{*}{Table~\ref{tab:depth_analysis}} & \multirow{4}{*}{\begin{tabular}[c]{@{}c@{}}Optimizer:\\      SGD\end{tabular}}            & Lr                   & 0.1            \\
                          &                                                                                           & Batch\_size          & 100            \\
                          &                                                                                           & Momentum             & 0.9            \\
                          &                                                                                           & Weight\_decay        & 0              \\ \cline{2-4} 
                          & \multirow{3}{*}{\begin{tabular}[c]{@{}c@{}}Lr\_scheduler:\\      MutiStepLR\end{tabular}} & Epoch                & 100            \\
                          &                                                                                           & Milestones           & {[}40,60,80{]} \\
                          &                                                                                           & beta                 & 0.2            \\ \cline{2-4} 
                          & \multirow{2}{*}{Data\_transformer}                                                        & RandomCrop(32)       & /              \\
                          &                                                                                           & Normalize            & /              \\ \hline
\multirow{9}{*}{Table~\ref{tab:alternatives}}  & \multirow{4}{*}{\begin{tabular}[c]{@{}c@{}}Optimizer:\\      SGD\end{tabular}}            & Lr                   & 0.1            \\
                          &                                                                                           & Batch\_size          & 100            \\
                          &                                                                                           & Momentum             & 0.9            \\
                          &                                                                                           & Weight\_decay        & 1.00E-04       \\ \cline{2-4} 
                          & \begin{tabular}[c]{@{}c@{}}Lr\_scheduler:\\      CosineAnnealingLR\end{tabular}           & Epoch                & 100            \\ \cline{2-4} 
                          & \multirow{4}{*}{Data\_transformer}                                                        & RandomCrop(32)       & /              \\
                          &                                                                                           & RandomRotation(15)   & /              \\
                          &                                                                                           & RandomHorizontalFlip & /              \\
                          &                                                                                           & Normalize            & /              \\ \hline
\end{tabular}
\end{table}

\subsection{ImageNet Training Details}
\label{appendix_imagenet}
\begin{table*}[htbp]
    \caption{Structures for ImageNet.\label{tab:ImageNet_structure}}
    \centering
    \begin{tabular}{c|c|c|c|c}
    \hline
    Stage & Output Size & ResNet-18   & ResNet-34  & ResNet-104  \\ \hline
    Conv1 & 112x112     & \multicolumn{3}{c}{7x7, 64, stride=2}  \\ \hline
    Conv2 & 56x56       & $\left[\begin{array}{c}\text{3x3, 64}\\ \text{3x3, 64}\end{array}\right]*2$            & $\left[\begin{array}{c}\text{3x3, 64}\\ \text{3x3, 64}\end{array}\right]*3$           & $\left[\begin{array}{c}\text{3x3, 64}\\ \text{3x3, 64}\end{array}\right]*3$            \\ \hline
    Conv3 & 28x28       & $\left[\begin{array}{c}\text{3x3, 128}\\ \text{3x3, 128}\end{array}\right]*2$            &$\left[\begin{array}{c}\text{3x3, 128}\\ \text{3x3, 128}\end{array}\right]*4$            & $\left[\begin{array}{c}\text{3x3, 128}\\ \text{3x3, 128}\end{array}\right]*8$            \\\hline
    Conv4 & 14x14       & $\left[\begin{array}{c}\text{3x3, 256}\\ \text{3x3, 256}\end{array}\right]*2$            &$\left[\begin{array}{c}\text{3x3, 256}\\ \text{3x3, 256}\end{array}\right]*6$            & $\left[\begin{array}{c}\text{3x3, 256}\\ \text{3x3, 256}\end{array}\right]*32$            \\\hline
    Conv5 & 7x7         & $\left[\begin{array}{c}\text{3x3, 512}\\ \text{3x3, 512}\end{array}\right]*2$            &$\left[\begin{array}{c}\text{3x3, 512}\\ \text{3x3, 512}\end{array}\right]*3$            & $\left[\begin{array}{c}\text{3x3, 512}\\ \text{3x3, 512}\end{array}\right]*8$            \\\hline
    FC    & 1x1         & \multicolumn{3}{c}{AveragePool, FC-1000} \\\hline
    \end{tabular}
\end{table*}
For the experiments on ImageNet~\cite{imagenet}, we mainly follow the architectures of the canonical ResNet as in Table~\ref{tab:ImageNet_structure}, but the original MaxPool after the stage of Conv1 is replaced by a stride-2 convolution at the beginning of the Conv2 stage.

The training recipe at first mainly follows that of He et al.~\cite{He_2016_resnet} for MS-ResNet18 and MS-ResNet34, i.e. a 224x224 random crop with horizontal flip for data augmentation and an SGD optimizer with a weight decay of 1e-4 and a momentum of 0.9. The training contains 125 epochs and the initial learning rate linearly increases by 0.1 for every 256 batchsize. We report the average test accuracy of the last five epochs as our results. 
\begin{table*}[]
    \centering
    \caption{Hyper-parameters for experiments on ImageNet.}
    \label{tab:my_label}
\begin{tabular}{cccc}
\hline
Experiment                      & Component                                                                       & Name                   & Value       \\ \hline
\multirow{8}{*}{MS-ResNet18/34} & \multirow{4}{*}{\begin{tabular}[c]{@{}c@{}}Optimizer:\\      SGD\end{tabular}}  & Lr                     & 0.1         \\
                                &                                                                                 & Batch\_size            & 256         \\
                                &                                                                                 & Momentum               & 0.9         \\
                                &                                                                                 & Weight\_decay          & 1e-4    \\ \cline{2-4} 
                                & \begin{tabular}[c]{@{}c@{}}Lr\_scheduler:\\      CosineAnnealingLR\end{tabular} & Epoch                  & 125         \\ \cline{2-4} 
                                & \multirow{3}{*}{Data\_Transformer}                                              & RandomResizedCrop(224) & /           \\
                                                                &                                                                                 & RandomHorizontalFlip              & 0.5           \\
                                &                                                                                 & Normalize              & /           \\\hline
\multirow{10}{*}{MS-ResNet104}  & \multirow{4}{*}{\begin{tabular}[c]{@{}c@{}}Optimizer:\\      SGD\end{tabular}}  & Lr                     & 0.05 (0.1)  \\
                                &                                                                                 & Batch\_size            & 256         \\
                                &                                                                                 & Momentum               & 0.9         \\
                                &                                                                                 & Weight\_decay          & 1e-5 (1e-4) \\ \cline{2-4} 
                                & \begin{tabular}[c]{@{}c@{}}Lr\_scheduler:\\      CosineAnnealingLR\end{tabular} & Epoch                  & 100 (200)    \\ \cline{2-4} 
                                & \multirow{3}{*}{Data\_Transformer}                                              & RandomResizedCrop(224) & /           \\
                                &                                                                                 & AutoAugment            & /           \\
                                &                                                                                 & Normalize              & /           \\ \cline{2-4} 
                                & CrossEntropyLoss                                                                & Label\_smoothing       & 0.1         \\\cline{2-4}
                                & Dropout                                                                         & p                      & 0.2         \\ \hline
                                \multicolumn{4}{l}{\footnotesize *The values in parentheses are for the pretrain stage of MS-ResNet104.}
\end{tabular}
\end{table*}
Unfortunately, we observe severe overfitting problem when extending the model to 104-layer deep. An accuracy increase of 15\% on the training set only leads to an increase of 0.5\% on the test set. Therefore, a more advanced training recipe is adopted to fully unleash the potential of the deep model.

For the training of MS-ResNet104, we use AutoAugment~\cite{Cubuk_2019_autoaugment} for stronger data augmentation, adopt label smoothing~\cite{Szegedy_2016_smoothlabel} and dropout~\cite{2012_dropout} before the fully-connected layer for stronger regularization. Besides, a 2x2 average pooling layer with a stride of 2 is added before each stride-2-CONV1x1 in the shortcut connection for downsampling, and the stride of the CONV1x1 is changed to 1 in order not to discard information~\cite{He_2019_bag_of_tricks}. 

A longer training procedure is expected to cooperate with the above techniques, but the direct training of SNNs requires time-consuming backpropagation through time. For saving training time, we propose a two-phase training process consisting of a T=1 pretrain phase and a formal training phase. The model is limited to one single timestep at first and gets trained for 200 epochs with large batchsize and learning rate, which is set to 0.1 per 256 batchsize. Then the pretrained model is utilized as a basis for further learning of temporal dynamics and efficient multi-timesteps expression. The finetune phase contains another 100 epochs and the model is trained with 1e-5 L2-penalty and the learning rate is set to 0.05 per 256 batchsize. In this way, we firstly obtain a T=1 model with an accuracy of 71.72\% and a T=6 model with an accuracy of 72.8\% if it is directly extended in the temporal dimension, and after the finetune stage, the model achieves an accuracy of 74.21\%. The T=1 pretrained model can provide a good starting point though it only occupies one-ten of the total training time, and the additional finetune phase further accommodates the model to effective spatiotemporal expression and gets superior performance. It is economical to use a T=1 SNN as the pretrained model to avoid time-consuming BPTT in SNNs with a large $T$.
    
The number of timesteps $T$ is set to 6 for training. In ResNet-104, however, we find that setting $T$ to 5 for only inference will cause no harm to the performance, so it is adopted in our final accuracy results and energy estimation.

The experiment is conducted on up to 8 NVIDIA RTX 2080Ti GPUs, depending on the memory cost. Synchronized BN technique will be used if multiple GPUs are required.

\bibliographystyle{IEEEtran}
\bibliography{bibliography.bib}


 




\vfill

\end{document}